\documentclass{article}

\usepackage{arxiv}

\usepackage[utf8]{inputenc} 
\usepackage[T1]{fontenc}    
\usepackage{hyperref}       
\usepackage{url}            
\usepackage{booktabs}       
\usepackage{amsfonts}       
\usepackage{nicefrac}       
\usepackage{microtype}      
\usepackage{lipsum}		
\usepackage{graphicx}
\usepackage[numbers]{natbib}
\usepackage{doi}
\usepackage{adjustbox}
\usepackage{pdflscape}

\title{Global Contrastive Training for Multimodal Electronic Health Records with Language Supervision}

\author{
  Yingbo Ma \\
  Department of Medicine \\
  University of Florida \\
  \texttt{yingbo.ma@ufl.edu} \\
  \And
  Suraj Kolla \\
  Department of Medicine \\
  University of Florida \\
  \texttt{n.kolla@ufl.edu} \\
  \And
  Zhenhong Hu \\
  Department of Medicine \\
  University of Florida \\
  \texttt{hzhuf@ufl.edu} \\
  \And
  Dhruv Kaliraman \\
  Department of Medicine \\
  University of Florida \\
  \texttt{dhruv.kaliram@ufl.edu} \\
  \And
  Victoria Nolan \\
  Department of Medicine \\
  University of Florida \\
  \texttt{vnolan@ufl.edu} \\
  \And
  Ziyuan Guan \\
  Department of Medicine \\
  University of Florida \\
  \texttt{ziyuan.guan@ufl.edu} \\
  \And
  Yuanfang Ren \\
  Department of Medicine \\
  University of Florida \\
  \texttt{renyuanfang@ufl.edu} \\
  \And
  Brooke Armfield \\
  Department of Medicine \\
  University of Florida \\
  \texttt{barmfield@ufl.edu} \\
  \And
  Tezcan Ozrazgat-Baslanti \\
  Department of Medicine \\
  University of Florida \\
  \texttt{tezcan@ufl.edu} \\
  \And
  Jeremy A. Balch \\
  Department of Surgery \\
  University of Florida \\
  \texttt{balch.jeremy@ufl.edu} \\
  \And
  Tyler J. Loftus \\
  Department of Surgery \\
  University of Florida \\
  \texttt{tloftus@ufl.edu} \\
  \And
  Parisa Rashidi \\
  Department of Biomedical Engineering \\
  University of Florida \\
  \texttt{parisa.rashidi@ufl.edu} \\
  \And
  Azra Bihorac$^{*}$ \\
  Department of Medicine \\
  University of Florida \\
  \texttt{abihorac@ufl.edu} \\
  \And
  Benjamin Shickel\thanks{Authors contributed equally} \\
  Department of Medicine \\
  University of Florida \\
  \texttt{shickelb@ufl.edu}
}

\begin{document}
\maketitle

\begin{abstract}
Modern electronic health records (EHRs) hold immense promise in tracking personalized patient health trajectories through sequential deep learning, owing to their extensive breadth, scale, and temporal granularity. Nonetheless, how to effectively leverage multiple modalities from EHRs poses significant challenges, given its complex characteristics such as high dimensionality, multimodality, sparsity, varied recording frequencies, and temporal irregularities. To this end, this paper introduces a novel multimodal contrastive learning framework, specifically focusing on medical time series and clinical notes. To tackle the challenge of sparsity and irregular time intervals in medical time series, the framework integrates temporal cross-attention transformers with a dynamic embedding and tokenization scheme for learning multimodal feature representations. To harness the interconnected relationships between medical time series and clinical notes, the framework equips a \textit{global contrastive loss}, aligning a patient's multimodal feature representations with the corresponding \textit{discharge summaries}. Since discharge summaries uniquely pertain to individual patients and represent a holistic view of the patient's hospital stay, machine learning models are led to learn discriminative multimodal features via global contrasting. Extensive experiments with a real-world EHR dataset demonstrated that our framework outperformed state-of-the-art approaches on the exemplar task of predicting the occurrence of nine postoperative complications for more than 120,000 major inpatient surgeries using multimodal data from UF health system split among three hospitals (UF Health Gainesville, UF Health Jacksonville, and UF Health Jacksonville-North).
\end{abstract}


\section{Introduction}
Electronic health records (EHRs) contain important information about patient encounters that support real-world healthcare delivery \citep{yadav2018mining}. While artificial intelligence and machine learning have the potential to support clinical decision-making based on contextual representations of patient data \citep{wijnberge2020effect}, modeling real-world EHRs remains challenging.

One of the challenges lies in modeling multivariate medical time series in EHRs, which inherently characterized by sparsity and irregular time intervals \citep{ghassemi2015multivariate} . Popular approaches such as recurrent neural networks (RNN) with long short-term memory (LSTM) \citep{memory2010long} and gated recurrent networks \citep{chung2014empirical} seek to account for the temporal complexities of medical time series, but may be suboptimal when learning long-term (e.g., over the duration of an entire hospital stay) temporal dynamics of patient health trajectories \citep{shickel2019deepsofa}. Recently, transformers have been used for modeling temporal EHR data \citep{li2020behrt,shickel2022multi,tipirneni2022self} and have been established as state-of-the-art approaches for predicting clinical outcomes from patient data sequences \citep{shickel2019deepsofa,balch2023building}. However, additional challenges persist when modeling EHR data with transformers, such as capturing temporal dependency across very long sequences \citep{li2022hi} and modeling heterogeneous dependencies across variables \citep{zhang2022crossformer}.

\begin{figure}[!htpb]
\centering 
\includegraphics[width=1\textwidth]{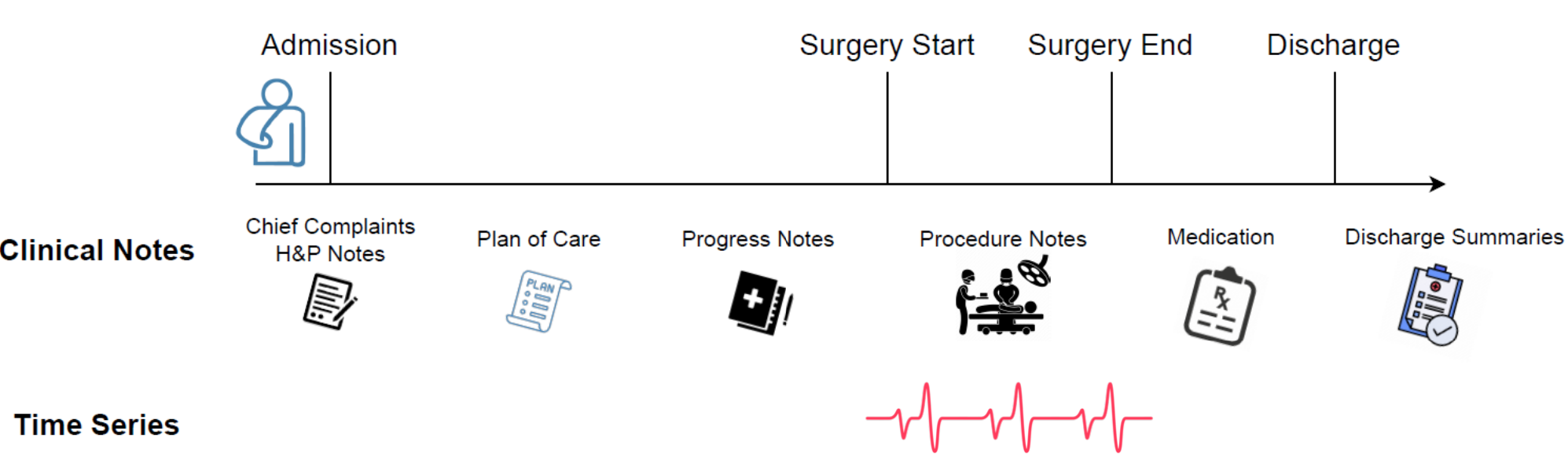} 
\caption{An example of unaligned, though clinically relevant, clinical notes and medical time series, where clinical notes are taken beginning patient admission through final discharge, including diverse and comprehensive information while medical time series only reflecting the patient's vital signs during a major surgery. This is ubiquitous in real-world EHR datasets.}
\label{fig:challenge} 
\end{figure}

Another challenge lies in effectively leveraging different modalities in EHRs and ensuring that their clinical aspects are meaningfully represented \citep{berger2016opportunities}. The complexity increases with the multimodal nature of EHR data, encompassing diverse clinical data from medical time series to radiology imaging data and unstructured clinical notes. Recent studies have shown the promise of using contrastive pre-training to jointly model different modalities for various multimodal clinical prediction tasks \citep{wang2022medclip, king2023multimodal}. Given two or more modalities, these contrastive pre-training methods generally learn bi-modal similarity scores, so that the data (e.g., chest radiographs and radiology report) from the same patient are pulled closer while those from different patients are pushed away. While effective, these methods may result in suboptimal learning when applied to real-world datasets, particularly when different modalities only capture specific perspectives of the patient's overall health trajectory \citep{yang2021leverage}. Therefore, in such scenarios, suboptimal performance maybe achieved when attempting to align two unimodal representations lacking sufficient shared information (See Figure~\ref{fig:challenge}).

To this end, we propose a novel global contrastive learning framework for multiomdal EHRs, specifically focusing on medical time series and clinical notes. To tackle the challenge of sparsity and irregular time intervals in medical time series, the framework integrates a dynamic embedding and tokenization scheme, using flexible positional encoding and a learnable time embedding to address the challenge of sparsity and irregular sampling, and a variable-specific encoding strategy for capturing distinct characteristics and relationships between temporal variables. To learn multimodal representations from unaligned medical time series and clinical notes, the framework equips a \textit{global contrastive loss}, aligning a patient's multimodal feature representations with the corresponding \textit{discharge summaries}. representing a holistic view of the patient's hospital stay. We demonstrate the effectiveness of our approach and analyze the relative contributions of each component of our framework using the benchmark task of predicting the onset of multiple postoperative complications following major inpatient surgery with a real-world EHR dataset .

\subsection*{Generalizable Insights about Machine Learning in the Context of Healthcare} 
This work makes several contributions to the ongoing research of exploring innovative approaches to incorporate diverse health data sources:
\begin{enumerate}
    \item Introducing a dynamic embedding and tokenization scheme with Longformer to enable transformers to adapt to the unique challenges of sparsity, multivariate, and irregular time intervals in multimodal medical time series.
    
    \item Introducing a novel global contrastive loss objective for multimodal EHRs to effectively leverage medical time series and clinical notes. Moreover, the framework is easier to scale up incorporating diverse modalities of health data, compared to the quadratic time and computational complexity of traditional approaches in which the contrastive loss between every two modalities needs to be calculated.

    \item Demonstrating the superior performance of the proposed framework with extensive experiments on a real-world EHR dataset collected from 113,953 adult patients who underwent 124,777 inpatient surgeries.
\end{enumerate}

\section{Related Work}
\subsection{Multimodal Representation Learning for Health}
Combining diverse sources of data sources in medical domain is promising for more comprehensive understanding of patients' health conditions \citep{raghupathi2014big}, more accurate health outcome predictions \citep{shickel2023dawn}, and building next-generation foundational medical models for generative AI \citep{moor2023foundation}. The core of this research effort is multimodal representation learning where all the modalities are projected to a common space while preserving information from the given modalities \citep{liang2022high}. Traditional data fusion methods, such as early concatenation fusion and late weighted average fusion \citep{xu2021mufasa}, are insufficient to learn the correlations and dependencies among different modalities \citep{ma2022detecting}. Recently, transformer-based architecture, thanks to its superior ability to capture cross-modal interactions by self-attention and its variants \citep{xu2023multimodal}, has achieved great success in various multimodal machine learning tasks in different domains, such as multimodal action recognition \citep{wang2020transmodality}, image segmentation \citep{xiao2023transformers}, and affect detection \citep{ma2023noisy}. Despite the promise, additional challenges persist when modeling multimodal EHR data with transformers, such as capturing temporal dependency across very long sequences \citep{li2022hi,ma2024temporal} and modeling heterogeneous dependencies across modalities \citep{zhang2022crossformer}.

\subsection{Contrastive Learning in Health}
Contrastive learning is a technique aimed at developing an embedding function capable of encoding input samples, potentially derived from various modalities, in a manner where samples originating from identical categories are proximally aligned, while those from disparate categories are distinctly separated within the embedding space. This approach supports both supervised and self-supervised learning paradigms, offering a versatile framework for data representation. There are a variety of loss functions in the literature that align with the contrastive training objective, starting with Contrastive Loss \citep{chopra2005learning}, which takes only pairs of samples from the input and minimizes embedding distance for the samples in the same class and maximizes the distance for samples in different classes. Triplet Loss \citep{schroff2015facenet} and N-pair Loss \citep{sohn2016improved} use the samples as anchors and select positive and negative samples with respect to them, minimizing and maximizing the distances between them, respectively. The triplet loss uses only one positive and negative sample per anchor, while the N-pair loss uses multiple negative samples with one positive sample. Contrastive learning is being used for better representing images \citep{chen2020simple}, texts \citep{gao2021simcse} and others by training the respective embeddings on the loss functions. It has also been widely adopted for multimodal representation learning by contrasting positive and negative pairs of instances \citep{le2020contrastive}.

In the healthcare domain, contrastive learning has demonstrated significant efficacy, particularly in tasks where labeled data is either scarce or entails high curation costs \citep{chaitanya2020contrastive}. This approach is not only beneficial in such data-constrained environments but also enhances supervised learning tasks. For instance, Azizi et al. \citep{azizi2021big} effectively applied contrastive learning for the pretraining of models, subsequently improving their performance in medical image classification tasks. The methodology gained further traction following its successful application in the CLIP model \citep{radford2021learning}, which adeptly aligns images with their textual captions. Its application has since expanded to the realm of multimodal EHRs, facilitating the alignment of disparate data forms, including chest radiographs with radiology reports  \citep{wang2022medclip, zhang2022contrastive}, medical time series with clinical notes \citep{king2023multimodal}, ICD codes with clinical notes \citep{koo2024next}, and the correlation of retinal images with genomic data \citep{taleb2022contig}, thereby illustrating the broad applicability of contrastive learning in complex data integration within healthcare.

\section{Methodology}
Consider a cohort $\mathbf{C}$ with multimodal EHRs, where $\mathbf{C}=\left ( p_i, \mathbf{S}_i, \mathbf{T}_i, \mathbf{N}_i, \mathbf{y}_i \right )$, in which for each patient $p_{i}$, the dataset contains the patient's static tabular data $\mathbf{S}_i$, such as demographic information, a multivariate medical time series $\mathbf{T}_i$, such as physiological vital signs, corresponding unstructured clinical notes $\mathbf{N}_i$, such as history and physical (H\&P) notes, and clinical outcomes $\mathbf{y}_i$, such as in-hospital mortality.

\begin{figure}[!htbp]
\centering 
\includegraphics[width=1\textwidth]{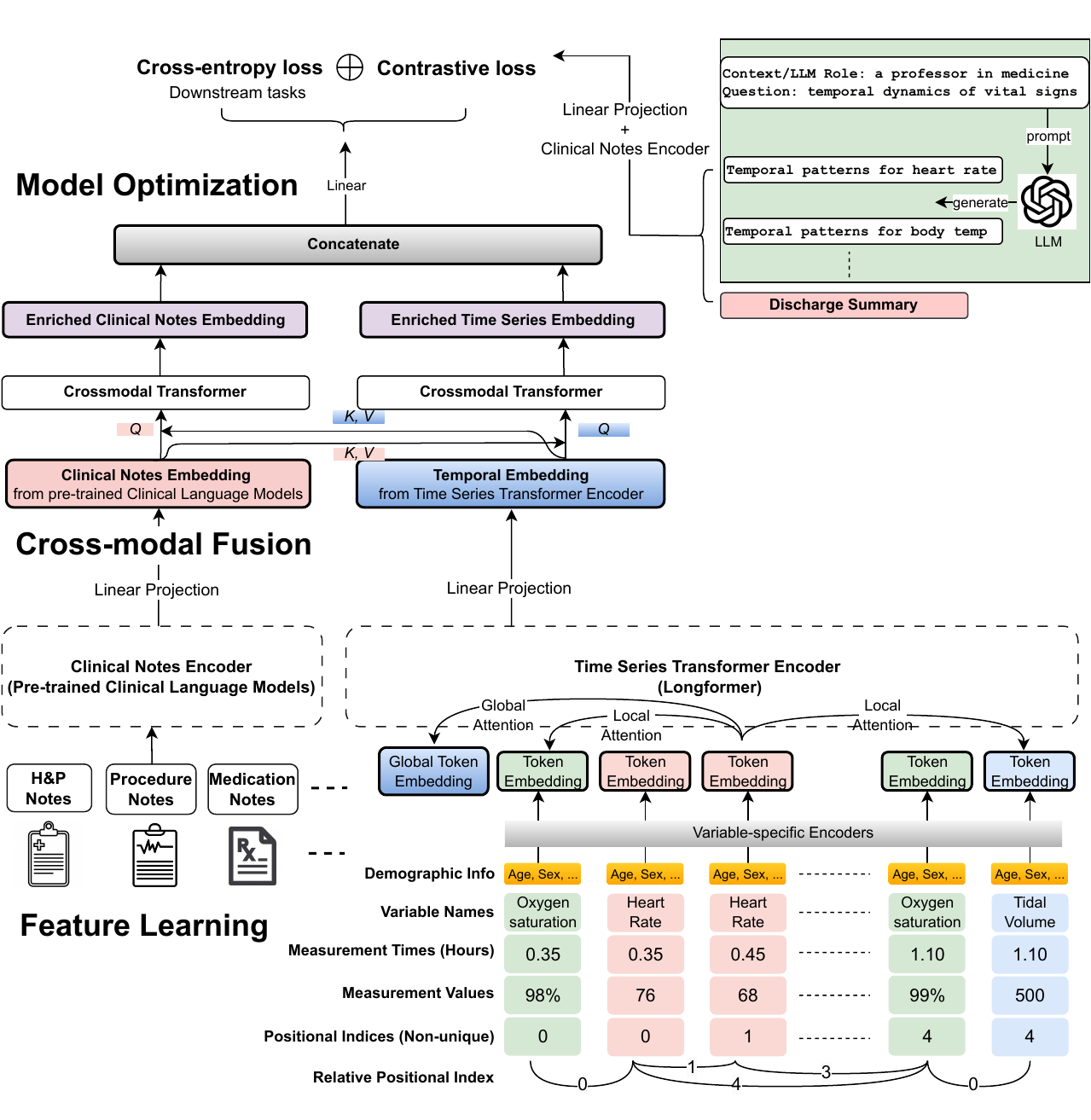} 
\caption{The overview of the proposed global contrastive learning framework for multimodal EHRs. The framework consists of three main components: (1) Feature learning sub-network extracts temporal representation from medical time series and textual representation from clinical notes. (2) cross-modal fusion sub-network merges unimodal features with transformers in the cross-attention fashion, in which feature embedding from one modality is enriched by searching for the
most relevant feature in the other modality; (3) modal optimization uses the learning objective combining both the cross-entropy loss between predictions and ground truth labels, and the contrastive loss aligning multimodal representations with discharge summaries improved by LLMs.}
\label{fig:overview} 
\end{figure}

An overview of our proposed framework for multimodal modeling of EHRs is shown in Figure~\ref{fig:overview}, which consists of three sub-networks: (1) multimodal feature learning, (2) cross-modal fusion, and (3) model optimization. Next, we will describe the details of each of these sub-networks.

\subsection{Multimodal Feature Learning}
The multimodal feature learning sub-network is designed to effectively learn the latent unimodal feature representation from each type of data in cohort $\mathbf{C}$, for medical time series $\mathbf{T}_i$ and clinical notes $\mathbf{N}_i$.

\subsubsection{Medical Time Series} \label{sec:3.1.1}
Learning useful representations of medical time series $\mathbf{T}_i$ is challenging due to its high dimensionality, sparsity, irregular and variable-specific recording frequency, and timestamp duplication when multiple measurements are recorded simultaneously. Popular approaches such as recurrent neural networks (RNN) with long short-term memory (LSTM) \citep{memory2010long} and gated recurrent networks \citep{chung2014empirical} do not account for the temporal complexities of EHR data and may be suboptimal when learning temporal dynamics of patient health trajectories. To tackle this challenge, we adopted and modified transformer-based models by introducing a dynamic embedding and tokenization scheme to enable transformers to adapt to the above-mentioned unique challenges of medical time series.

\paragraph{Flexible positional encoding:} Multivariate EHR time series contain variables measured at different frequencies. To adapt to this unique challenge, we propose to use non-unique absolute positional indices based on the recorded timestamps so that variable tokens measured at the same time will be assigned the same positional index; in addition, we add a relative positional encoding to each token embedding \citep{shaw2018self}, which can help capture local token dependencies, especially for processing long sequences\citep{zaheer2020big,wei2021position}, to model the relationships between clusters of short-term activity across a long timeframe.

\paragraph{Learnable time encoding:} Positional embeddings alone omit critical information about the relative time between events. For applications of transformers to time series, time embeddings can help capture important temporal patterns. We propose to use Time2Vec \citep{kazemi2019time2vec} to learn a model-agnostic vector representation for time. In Time2Vec, a time \textit{t} is encoded to generate one non-periodic $\omega_{np}  t + \phi_{np} $, and one periodic $sin(\omega_{p}  t + \phi_{p})$ time dependent vector, where $\omega$ and $\phi$ are learnable parameters \citep{liang2023learn}.

\paragraph{Variable-specific encoding:} A multivariate clinical time series often includes different categories of health variables (e.g., vital signs, laboratory tests, medications) that tend to exhibit distinct characteristics, numerical ranges, and temporal patterns. To learn the unique characteristics of each time series, we propose to use a separate encoder for each clinical variable for \textit{intra-variable} temporal dynamics, and then concatenate the outputs of the separate encoders to learn the \textit{inter-variable} correlation and dependencies.

We embedded patient's static tabular data $\mathbf{S}_i$ and aggregated into each token in the sequence as a default contextualized information. We extended the notion of ``CLS'' token and prepended the global token to the tokenized sequence. The global token self-attended to all sequence elements, and was used to generate the latent feature representation of medical time series $\mathbf{T}_i$.

\subsubsection{Clinical Notes}
We encoded clinical notes with pre-trained clinical large language models. Clinical large language models are pre-trained to learn generally useful representations from the knowledge encoded in medical corpora, and have shown state-of-the-art performance in medical NLP tasks such as medical information extraction and question answering \citep{singhal2023large}. In this work, we experimented with popular clinical large language models including BioBERT \citep{lee2020biobert}, Clinical BERT \citep{alsentzer2019publicly}, Clinical Longformer \citep{li2022clinical}, and GatorTron \citep{yang2022large}, and selected the one which yielded the best performance.

The outputs from time series transformer encoder $X_{time}$ and clinical notes encoder $X_{note}$ are later passed through a linear projection layer respectively. This step maps each encoder’s representation to the multimodal embedding space, preparing the inputs for the next cross-modal fusion sub-network.

\subsection{Cross-modal Fusion}
To learn multimodal representations, we merged the embeddings of medical time series $X_{time}$ and clinical notes $X_{note}$ using a validated cross-attention-based approach \citep{tsai2019multimodal}, in which each feature embedding in one modality is enriched by searching for the most relevant feature in the other modality. For example,
\begin{equation}
    W_{note \rightarrow time} = softmax(\frac{Q_{time}K_{note}^{T}}{\sqrt{d_{k}}})
\end{equation} 
represent a scoring matrix, whose $(i, j)$-th element measures the attention given by the information from the $i$-th time step from modality $X_{time}$ and the $j$-th time step from modality $X_{note}$. Then, $\hat{X_{time}}$, the final enriched feature sequence for $X_{time}$ as:
\begin{equation}
    \hat{X_{time}}=W_{note \rightarrow time}V_{note}
\end{equation}
where $Q$, $K$, and $V$ denote Query, Key, and Value, respectively. Finally, the enriched feature sequence $\hat{X_{time}}$ and $\hat{X_{note}}$ are concatenated and projected to lower feature space as $h_{time+note}$ for downstream classification or prediction tasks with linear layers:
\begin{equation}
    h_{time+note}=Linear(concat(\hat{X_{time}}; \hat{X_{note}}))
\end{equation}

\subsection{Contrasting Multimodal Representation and Discharge Summaries} \label{sec:3.3}
Previous research in the medical field employing contrastive learning has yielded encouraging outcomes, particularly in contrasting cross-modal data such as medical images and texts \citep{wang2022medclip}. A key factor contributing to the effective joint training of medical images and texts lies in the significant \textit{shared semantics} observed between chest radiographs and their corresponding radiology reports \citep{li2023unify}, with the latter serving as explicit textual representations of the former. However, the application of contrastive learning to medical time series and clinical notes presents a distinct challenge. The semantic relationship between these modalities is not necessarily one of direct \textit{shared} meaning but rather of a \textit{complementary} nature, with each modality offering a unique perspective on the patient's health trajectory, insights not seamlessly interchangeable between the modalities, underscoring the complexity of their semantic relationship \citep{liang2022foundations}. Medical time series, typically recorded in intensive care units (ICU), provide a snapshot of a patient's vital signs over a brief interval. In contrast, clinical notes span the entirety of a patient's hospital stay, from admission to discharge, encompassing a diverse array of information, including \textit{surgery procedures} and \textit{medications}. Although both modalities hold clinical significance, their disparate nature may result in limited shared information, potentially hindering the efficacy of inter-modality alignment and leading to suboptimal learning outcomes when employing a straightforward contrasting approach.

To this end, we propose to use the multimodal latent representation $h_{time+note}$ from an individual patient $p_i$ to pair with the patient's corresponding \textit{discharge summary} as the contrastive objective. Employing multimodal representations for contrastive learning is recognized for its efficacy in capturing a global perspective of semantic information within patient health records \citep{mai2023learning}. An essential step is to define appropriate positive and negative samples for each patient's multimodal representation $h_{time+note}$. The rationale for selecting discharge summaries lies in their comprehensive nature, encapsulating an all-encompassing overview of a patient’s entire hospitalization \citep{cui2024multimodal}. Discharge summaries are particularly rich in information, often encompassing patient demographics, social context, details of admission, physical exam findings, diagnoses, and specifics of any therapeutic or diagnostic procedures undertaken, along with prescribed medications and the summary of hospital course \citep{ellershaw2024automated}. This wealth of information aligns well with the varied data modalities present in EHRs. Moreover, discharge summaries are commonly excluded from the development of predictive models using multimodal EHRs, primarily due to concerns about data leakage, particularly as they often contain conclusive assessments of patients’ critical conditions \citep{li2020inferring}. This exclusion, however, positions them as a potentially advantageous addition for learning objectives during model training. Finally, since each patient has a unique corresponding discharge summary, in doing so, we can easily set the positive pairs to the batch size $K$ during training. We define our contrastive learning objective as the addition of two asymmetric losses: 
\begin{equation}
    \resizebox{.25\hsize}{!}{$\mathcal{L}_{alignment} = \mathcal{L}_{MD} + \mathcal{L}_{DM}$}
\end{equation}
where $\mathcal{L}_{MD}$ denotes the loss contrasting multimodal representation with discharge summaries, and $\mathcal{L}_{DM}$ denotes the loss contrasting discharge summaries with multimodal representation. $\mathcal{L}_{MD}$ is calculated as:
\begin{equation}
    \resizebox{.55\hsize}{!}{$\mathcal{L}_{MD}=-\frac{1}{2K} \sum_{i=1}^N (\log \frac{\exp \left(\left\langle\hat{\mathbf{H}}_i^M, \hat{\mathbf{H}}_i^D\right\rangle / \tau\right)}{\sum_{\hat{\mathbf{H}}^{-D} \in \mathcal{N}^D} \exp \left(\left\langle\hat{\mathbf{H}}_i^M, \hat{\mathbf{H}}^{-D}\right\rangle / \tau\right)})$}
\end{equation}
and $\mathcal{L}_{DM}$ is calculated as:
\begin{equation}
    \resizebox{.55\hsize}{!}{$\mathcal{L}_{DM}=-\frac{1}{2K} \sum_{j=1}^N (\log \frac{\exp \left(\left\langle\hat{\mathbf{H}}_j^M, \hat{\mathbf{H}}_j^D\right\rangle / \tau\right)}{\sum_{\hat{\mathbf{H}}^{-M} \in \mathcal{N}^M} \exp \left(\left\langle\hat{\mathbf{H}}^{-M}, \hat{\mathbf{H}_j}^D\right\rangle / \tau\right)})$}
\end{equation}
Here $\left\langle,\right\rangle$ is cosine similarity, $\tau$ is the temperature hyperparameter modulating distribution’s concentration and Softmax function’s gradient, and $K$ is the batch size. By minimizing this loss, the similarity between each patient's multimodal representation and the corresponding discharge summaries increased while the similarity between the multimodal representation and any other patients' discharge summaries are decreased.

The final loss function for model optimization $\mathcal{L}_{total}$ is a weighted sum of the contrastive loss $\mathcal{L}_{alignment}$ and the cross-entropy loss $\mathcal{L}_{ce}$ for any downstream tasks between ground truth and prediction:
\begin{equation}
    \mathcal{L}_{total} = \alpha \mathcal{L}_{alignment} + \beta \mathcal{L}_{ce}
\end{equation}
where $\alpha$ and $\beta$ are parameters that balance the different loss terms \citep{koo2024next}.

\subsection{Improving Discharge Summaries with LLMs}
Discharge summaries represent a holistic view of a patient's visit beginning admission through discharge, hence corresponds well to the patient's clinical notes, it may not correspond to medical time series well. Medical time series are collected in high-frequency and the hidden temporal patterns from medical time series are important indicators to a patient's health trajectory \citep{ren2022performance}. Yet, discharge summaries often omit the textual description for medical time series. Hence, this could lead to suboptimal contrastive learning performance when aligning multimodal representation $h_{time+note}$ with discharge summaries. 

\begin{figure}[t]
\centering 
\includegraphics[width=.85\textwidth]{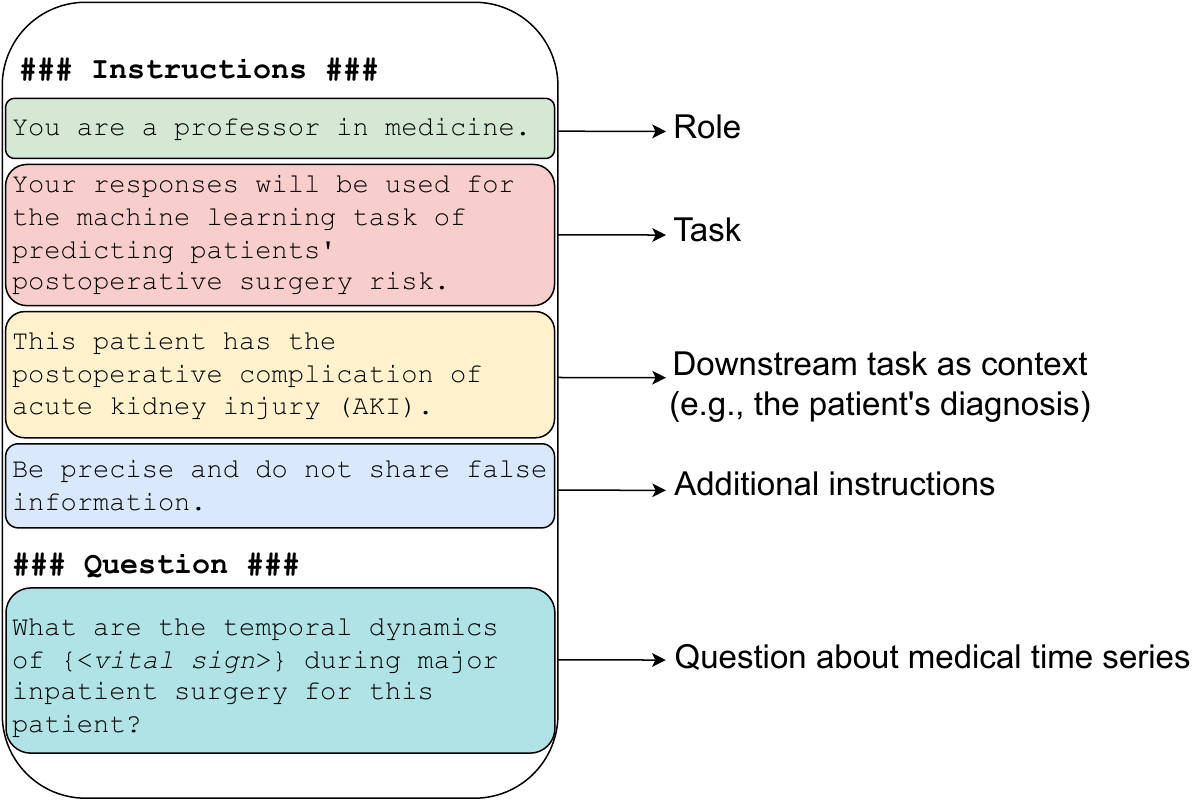} 
\caption{Prompting LLMs generating in-context texts for medical time series.}
\label{fig:prompts} 
\end{figure}

To better align the textual semantics of discharge notes with multimodal representation $h_{time+note}$, we propose to further improve discharge summaries with additional zero-shot LLM-generated textual descriptions for medical time series. An example of prompting templates is shown in Figure ~\ref{fig:prompts} 
. In this example, we prompted an LLM to answer the question with regard to a specific medical time series, with the role of LLM set as ``a professor in medicine''. We also provided the LLM with the downstream learning task (in this case, ``predicting postoperative surgery risk using intraoperative medical time series'') as a general context, and the ground truth for a patient in the learning task as an individualized context (in this case, ``AKI''). Additional instructions were included to alleviate hallucination. Finally, we prompted the LLM with the question regarding the temporal patterns of a specific physiological vital sign.

After generating the textual description for medical time series from LLM, we combined it with the patient's discharge summaries, and utilized the improved discharge summaries for each patient to contrast with the patient's multimodal representation $h_{time+note}$. With the additional inserted texts regarding medical time series, the improved discharge summaries are designed to be more contextualized matched with the patient's multimodal representation $h_{time+note}$ compared to the original discharge summaries.

\section{Cohort Selection and Data Preprocessing}
In this section, we describe the retrospective dataset for evaluating our approach on the benchmark task of predicting multiple in-hospital complications of major inpatient surgery using a real-world EHR dataset.

Our dataset consists of complete EHR records for all major inpatient surgeries occurring at three medical centers (UF Health Gainseville, UF Health Jacksonville, and UF Health North Jacksonville) between 2014 and 2019. The combined cohort consisted of 113,953 adult patients who underwent 124,777 inpatient surgeries. When a patient had multiple surgeries during one admission, only the first surgery was included. For each inpatient surgery, our dataset consists of:
\begin{enumerate}
    \item 9 preoperative demographic and admission information from 113,953 patients, including age (Mean 51 y, Min 18 y, Max 106 y), sex (48\% male, 52\% female), language, ethnicity, race, smoking status, zip code, and body mass index.

    \item 14 intraoperative temporal vital signs, including systolic blood pressure, diastolic blood pressure, mean arterial pressure, heart rate, respiratory rate, oxygen flow rate, fraction of inspired oxygen (FIO2), oxygen saturation (SPO2), end-tidal carbon dioxide (ETCO2), minimum alveolar concentration (MAC), positive end-expiratory pressure (PEEP), peak inspiratory pressure (PIP), tidal volume, and body temperature.

    \item 173 types of all preoperative and intraopertive clinical notes for an encounter, such as History and Physical (H\&P notes) and operative reports.

    \item 9 major postoperative complications \citep{ren2022performance} as binary clinical outcomes, the incidence of complications include 23.29\% ICU stay (for 48 h or more), 13.09\% acute kidney injury, 8.64\% prolonged mechanical ventilation, 2.00\% in-hospital mortality, 13.48\% wound complications, 15.09\% neurological complications, 8.20\% sepsis, 12.18\% cardiovascular complications, and 4.51\% venous thromboembolism.
\end{enumerate}

For demographic and admission information, we converted each to one-hot vectors, and concatenated with remaining numerical values. Missing static features was imputed with cohort medians.

For 14 intraoperative time series data, their variable names were converted to unique integer identifiers; the measured values for each variable were normalized to zero mean and unit variance based on the values from the training set; their measurement time, in the format of ``month/day/year hour:min:sec'', were first converted to unix timestamps and then also normalized similarly. For absolute positional indices, we assign one integer positional index for each token yet not enforcing the restriction that positional indices are unique and if different variables were measured at the same time. For relative positional embeddings, we generated the relative positional representation based on the GitHub code for the original paper \citep{shaw2018self}. The maximum sequence length of tokens in our dataset is 14,126, and the mean sequence length is 2,023.

For clinical notes, in the preprocessing phase, we merged all types of notes per surgery, converted the text to lowercase, and removed special characters and de-identification placeholders. Subsequently, we generated embeddings by first tokenizing the whole text using the clinically pretrained tokenizer. The tokens were then chunked to fit the pretrained clinical LLM, and the last hidden layer output for the CLS token was extracted as the embedding for each chunk. The final representation for each surgery was obtained by calculating the average of all these embeddings. We fixated on the Clinical Longformer \citep{li2022clinical} for generating the embeddings due to its superior performance in classifying with clinical notes, following extensive testing with various models from Huggingface including BioBERT \citep{lee2020biobert}, BiomedBERT \citep{chakraborty2020biomedbert}, ClinicalBERT \citep{alsentzer2019publicly}, Clinical Longformer \citep{li2022clinical}, and GatorTron \citep{yang2022large}.

\section{Experiments}
\subsection{Benchmark Multitask Classification}
The goal is to predict the onset of nine postoperative complications following major inpatient surgery: prolonged ($>$ 48 hours) intensive care unit (ICU) stay, acute kidney injury (AKI), prolonged mechanical ventilation (MV), wound complications, neurological complications, sepsis, cardiovascular complications, venous thromboembolism (VTE), and in-hospital mortality. Models are trained on data available in the EHR up to the recorded surgery end timestamp.

Our model was trained with the multi-task fashion for predicting 9 postoperative outcomes. To do this, we expanded the notion of ``[CLS]'' token for text classification and prepended 9 global tokens to our tokenized sequences, one for each of our postoperative outcomes, so that self-attentions were computed among all sequence elements for each clinical outcome token.

\subsection{Experimental Setup}
We used the following hyperparameters for optimization and regularization, Adam optimizer with a learning rate of 1e-4, dropout of 0.2, and weight decay of 1e-4. For the transformer models, including the Longformer, we limited the models to only 1 attention head and 1 layer per head, as this configuration produced the best results. We trained the models on two NVIDIA A100-SXM4-80GB GPUs for 30 epochs to leverage hardware acceleration. We used a batch size of 32 per GPU for the best performing model.

\subsection{Examining the Performance of Modeling Medical Time Series}
In this paper, we first proposed a novel dynamic embedding and tokenization scheme to modeling medical time series, which introduces three novelties to existing approaches: a flexible positional encoding, a learnable time encoding, and variable-specific encoding. In this work, we experimented with three types of variable-specific encoders, including 1-D convolutional encoder \citep{kiranyaz20211d}, transformer encoder \citep{tipirneni2022self}, and linear encoder. This subsection examines the performance of modeling clinical time series by comparing the model trained with the dynamic tokenization scheme introduced in this paper with a few widely adopted baselines: (1) Tokenized gated recurrent units (GRUs) with attention: GRU is a popular sequential network for tasks involving modeling clinical time series \citep{tan2020data,shi2021deep}. (2) Tokenized XGBoost: XGBoost gradient boosting algorithm employs gradient boosting on decision trees for regression and classiﬁcation tasks \citep{wang2020utilizing,liu2022dynamic}. (3) BEHRT \citep{li2020behrt}: BEHRT is a transformer-based model with traditional tokenization scheme, a widely used baseline transformer model for EHR data. (4): Hi-BEHRT \citep{li2022hi}: Hi-BEHRT is a hierarchical transformer model extending from BEHRT, specifically designed for processing longer EHR sequences. Hi-BEHRT uses a sliding window to segment the full sequence into smaller segments and applied transformers as local feature extractor for the temporal interaction within each segment. (5) Self-supervised Transformer for Time-Series (STraTS) \citep{tipirneni2022self}: STraTS uses a unique transformer to encode each variable and then uses a self-attention layer to generate the time-series embedding.

\subsection{Examining the Performance of Multimodal Contrastive Learning}
Another novelty of this work is the contrastive loss between multimodal representation and improved discharge summaries by LLMs. Despite the success of multimodal contrastive learning between medical images and clinical notes (e.g., MedCLIP \citep{wang2022medclip}), extending this to the alignment between medical time series and clinical notes remains challenging, due to the potential lack of shared semantic information between these two modalities as described in Subsection~\ref{sec:3.3}. In this paper, we took a different approach and proposed to align multimodal representation with the semantics of corresponding discharge summaries. We compared the performance of the models trained with our approach with the baseline model incorporating the inter-modality contrastive loss. The inter-modality contrastive loss calculates the bi-modal contrastive loss pairing each two modalities of data in EHRs during the training process. This baseline approach extends the CLIP model \citep{radford2021learning} and have been widely adopted on the task of pre-training between medical time series and clinical notes \citep{king2023multimodal} and between medical images and clinical notes. We also trained the model only using the ground truth labels in a supervised fashion without the contrastive loss.

\begin{landscape}
\begin{table}[!htbp]
\centering
\caption{Examining the performance of modeling medical time series. Comparing the AUROC scores of different approaches for predicting nine postoperative outcomes.}
\label{tab:time_modeling_auroc}
\begin{adjustbox}{width = 1.4\textwidth}
\begin{tabular}{@{}ccccccccccc@{}}
\toprule
Model                & Mean  & ICU   & AKI   & MV    & Mortality & Wound & Neurological & Sepsis & Cardiovascular & VTE   \\ \midrule
GRU + Attention \citep{shi2021deep}     & 0.771$\pm 0.04$  & 0.857 & 0.718 & 0.783 & 0.816 & 0.712 & 0.753 & 0.791 & 0.762 & 0.747 \\
XGBoost \citep{liu2022dynamic}             & 0.765$\pm 0.03$ & 0.851 & 0.716 & 0.771 & 0.815 & 0.709 & 0.748 & 0.788 & 0.760 & 0.727 \\
BEHRT \citep{li2020behrt} & 0.749$\pm 0.01$ & 0.843 & 0.701 & 0.765 & 0.800 & 0.701 & 0.725 & 0.770 & 0.748 & 0.699      \\
Hi-BEHRT \citep{li2022hi}             & 0.781$\pm 0.03$ & 0.863 & 0.730 & 0.789 & 0.835 & 0.721 & 0.769        & 0.801 & 0.780          & 0.769 \\
STraTS \citep{tipirneni2022self}            & 0.797$\pm 0.04$ & 0.881 & 0.742 & 0.803 & \textbf{0.857} & 0.734 & 0.797        & \textbf{0.813} & 0.791          & 0.772 \\ \midrule
Longformer + Single-shared Encoder  & 0.780$\pm 0.03$ & 0.860 & 0.731 & 0.787 & 0.832 & 0.718 & 0.765 & 0.798  & 0.781 & 0.766 \\
Longformer + Variable-specific Encoder (1-D CNN)  & 0.796$\pm 0.02$ & 0.880 & 0.743 & 0.797 & 0.850 & 0.734 & 0.794 & 0.808  & 0.793 & 0.771 \\
Longformer + Variable-specific Encoder (Transformer)  & 0.800$\pm 0.03$ & 0.882 & 0.745 & 0.806 & 0.854 & 0.736 & 0.798 & 0.810 & 0.795 & 0.772 \\
Longformer + Variable-specific Encoder (Linear)  & 0.798$\pm 0.02$ & 0.880 & 0.743 & 0.803 & 0.853 & 0.735 & 0.796 & 0.808 & 0.793 & 0.771 \\
Longformer + Variable-specific Encoder (Linear) + Time Embedding  & \textbf{0.801$\pm 0.02$} & \textbf{0.883} & \textbf{0.749} & \textbf{0.810} & 0.853 & \textbf{0.739} & \textbf{0.800} & 0.811  & \textbf{0.797} & \textbf{0.774} \\ \bottomrule
\end{tabular}
\end{adjustbox}
\end{table}

\begin{table}[!htbp]
\centering
\caption{Examining the performance of multimodal contrastive learning. Shown are the AUROC of different approaches for predicting nine postoperative outcomes.}
\label{tab:contrastive_modeling_auroc}
\begin{adjustbox}{width = 1.4\textwidth}
\begin{tabular}{@{}ccccccccccc@{}}
\toprule
Training Loss & Mean & ICU & AKI & MV & Mortality & Wound & Neurological & Sepsis & Cardiovascular & VTE \\ \midrule
Cross-entropy loss (time series + notes $\leftrightarrow$ ground-truth) \citep{shickel2023dynamic} & 0.845$\pm 0.03$ & 0.908 & 0.781 & 0.845 & 0.905 & 0.780 & 0.855 & 0.882 & 0.823 & 0.838 \\ 
Inter-modality contrastive loss (time series $\leftrightarrow$ notes ) \citep{king2023multimodal} & 0.738$\pm 0.02$ & 0.812 & 0.687 & 0.756 & 0.813 & 0.660 & 0.750 & 0.795 & 0.722 & 0.736 \\
Cross-entropy loss + Inter-modality contrastive loss \citep{koo2024next} & 0.843$\pm 0.03$ & 0.908 & 0.780 & 0.843 & 0.901 & 0.780 & 0.850 & 0.880 & 0.818 & 0.836 \\ \midrule
Multimodal contrastive loss (time series + notes $\leftrightarrow$ discharge summaries) & 0.803$\pm 0.04$ & 0.877 & 0.752 & 0.821 & 0.878 & 0.725 & 0.815 & 0.860 & 0.787 & 0.801 \\
Cross-entropy loss + Multimodal contrastive loss & 0.859$\pm 0.04$ & 0.916 & 0.801 & 0.855 & 0.915 & 0.789 & 0.871 & 0.894 & 0.831 & 0.849 \\
Cross-entropy loss + Multimodal contrastive loss (LLM-improved) & \textbf{0.862}$\pm 0.03$ & \textbf{0.918} & \textbf{0.805} & \textbf{0.857} & \textbf{0.918} & \textbf{0.796} & \textbf{0.875} & \textbf{0.904} & \textbf{0.834} & \textbf{0.851} \\
\bottomrule
\end{tabular}
\end{adjustbox}
\end{table}
\end{landscape}

\section{Results and Discussion}
This paper proposed a contrastive learning framework for modeling multimodal EHRs, specifically focusing on medical time series and clinical notes. To tackle the challenge of modeling medical time series, we modified Longformer and introduced a dynamic embedding and tokenization scheme. For improving the multimodal fusion of medical time series and clinical notes, we proposed to use the multimodal contrastive loss (medical time series + clinical notes) with discharge summaries, instead of aligning between these two modalities. This section summarizes and discusses the experimental results for the framework.

\subsection{The Dynamic Tokenization and Embedding Scheme for Time Series}
Table~\ref{tab:time_modeling_auroc} compares the area under receiver operating characteristic (AUROC) scores of our proposed Longformer-based approach with several popular baselines for time series modeling. As shown in the table, our dynamic tokenization scheme-based Longformer model with additional variable-specific encoders and time embedding outperformed all baseline models with the highest mean AUROC of 0.801. STraTS \citep{tipirneni2022self} slightly under-performed our approach, suggesting the effectiveness of incorporating extra relative positional embeddings in our approach. With the same tokenized sequence, GRU + Attention (AUROC: 0.771) performed better than transformer models with traditional tokenization scheme (AUROC: 0.749), indicating the superior performance from transformer-based models for this task.

We further ablated the variable-specific encoder and time embedding to examine the benefits brought by each of these components. For encoding each variable separately, as described in Subsection~\ref{sec:3.1.1}, we introduced variable-specific encoders to Longformer for capturing temporal dynamics within each variable, instead of using a single-shared encoder. Table~\ref{tab:time_modeling_auroc} reported the performance of using Longformer with the default single-shared encoder against Longformer with three different types of variable-specific encoders (1-D convolutional layers \citep{kiranyaz20211d} and transformer layers \citep{tipirneni2022self}). The results showed the effectiveness of using a variable-specific encoder indicated by the performance lift, and the model performance of using different types of encoders were similar (For 1-D CNN, transformer, and linear, the mean AUROC score was 0.796, 0.800, and 0.798, respectively). Considering the time complexity of 1-D CNN and transformer encoders as well as generalizing this approach to high-counts of temporal variables, we chose linear encoder for the experiments reported in the rest of the paper. For examining the effectiveness of time embedding, experimental results of adding additional time embedding further improved the AUROC by 0.03.

\subsection{Intermodal Contrasting versus Multimodal Contrasting}
Table~\ref{tab:contrastive_modeling_auroc} compares the AUROC scores of our proposed multimodal contrastive learning approach with several baseline training objectives. As shown in the table, we first experimented several baseline models trained with the cross-entropy loss (between ground truth postoperative outcomes and prediction) and inter-modality contrastive loss. The models only trained with inter-modality contrastive loss yielded the poor AUROC score of 0.738, indicating low extent of shared semantic information between medical time series and clinical notes in our dataset. In addition, the model trained with the combined loss of cross-entropy and inter-modality contrasting (AUROC of 0.843) did not outperform the model only trained with the cross-entropy loss (AUROC of 0.845).

The models trained with our approach (multimodal representation contrasting to dis- charge summaries) achieved better performance. The models only trained with the proposed multimodal contrastive loss achieved the AUROC score of 0.803; the model trained with the combined loss of cross-entropy and multimodal contrasting outperformed all baseline models with the AUROC score of 0.859. The models trained with the LLM-improved discharge summaries performed the best, establishing a state-of-the-art mean AUROC of 0.862 for this task.

Although the traditional intermodal contrastive learning is successful in benchmarking public datasets such as MIMIC-III \citep{johnson2016mimic}, it achieved a relatively poor performance on our dataset, where medical time series were patients’ physiological vitals signs measured during surgery and clinical notes were taken throughout the encounter. In our dataset, the textual descriptions of vital signs appeared only in progress notes, in which the most recent values were auto-populated. Compared to the time series with high-frequency recording, textual descriptions do not provide useful information related to the temporal patterns. Therefore, these two modalities of data lack enough shared information for machine learning models to align, although they were both clinically relevant.

This problem has also been discussed in recent literature, as intermodal contrasting based on unimodal representation could be inaccurate, failing to reveal the global information in multimodal objects \citep{mai2023learning}. Compared to the traditional inter-modal contrastive learning, our framework took a different approach by using patients’ discharge summaries as a global supervision, directly aligning with multimodal representation combining medical time series and clinical notes, establishing the state-of-the-art performance on this task. In addition, our approach is easy to scale up, incorporating diverse modalities of health data, as intermodal contrastive learning requires quadratic time complexity as the number of modalities in EHRs increases.

\subsection{Improving Discharge Notes with LLMs}
Considering the scarcity of textual descriptions about medical time series in discharge summaries, our framework used LLMs to generate relevant texts describing these temporal dynamics. Similar LLM-based text augmentation approaches have also been successfully studied in other tasks, such as in-context rewriting for more robust sentiment analysis \citep{o2024improving}. 

In this study, we prompted ``{\fontfamily{qcr}\selectfont gpt-4-1106-preview}'' to generate the temporal patterns for each of the vital signs in our dataset, using patients’ diagnoses as context. This was designed for adding more contextual information to medical time series by inserting texts related to temporal patterns of medical time series during the model training process, in which machine learning models were trained to align the multimodal representation from both medical time series and clinical notes with discharge summaries. By improving the discharge summaries, the experimental results showed a slight performance improvement.

\subsection{Limitations and Future Work}
This work has several important limitations. First, the global contrastive learning framework was designed for the challenge of information unalignment in multimodal EHRs, a common problem with medical time series and clinical notes in real-world EHR datasets. It has not yet been applied to other modalities in EHRs, such as medical images, medication orders, and lab results. Future work needs to investigate the generalizability of the proposed framework on other datasets with more modalities of EHR data. Second, the framework is not suitable for online applications in prospective, in-patient early prediction tasks since it requires discharge summaries as the contrasting learning objective. A prospective dataset would likely be adversely affected by backlogging of vital signs and delays in documentation (e.g., for examples, procedure notes for inpatient surgery are typically written after surgery end time). Nevertheless, our experiments serve as a proof-of-concept for using clinical text as a contrastive medium for multimodal datasets. Third, discharge summaries themselves are not necessarily the ground truth for clinical events during a hospital stay and suffer from many of the limitations of clinical notes, including incompleteness and factual inaccuracies. Finally, the prompting methods for improving discharge summaries with LLMs was relatively simple. Future work needs to investigate more effective prompting techniques for improving discharge summaries.

\section{Conclusion}
EHRs offer potential for tracking personalized patient health trajectories using deep learning, but face challenges due to EHR data being characterized by high dimensionality, sparsity, irregular time intervals, and multiple modalities. Traditional contrastive pre-training methods have shown promise in jointly modeling multiple modalities for clinical prediction tasks, but they may be sub-optimal when the modalities represent specific perspectives of a patient’s overall health trajectory, lacking enough shared information for alignment. To address these challenges, this paper introduces a global contrastive learning framework for multimodal electronic health records using temporal cross-attention transformers with a dynamic embedding and tokenization scheme , and a global contrastive loss to align a patient’s multimodal feature representations to discharge summaries. Using a real-world dataset with multimodal data collected to predict postoperative complications, our framework established a new state-of-the-art performance.

\section*{Acknowledgement}
We would like to thank the NVIDIA Corporation for providing computational resources used for this research. This research was also supported by National Institute of Health (NIH) through grant R01 GM110240. Any opinions, findings, conclusions, or recommendations expressed in this research are those of the authors, and do not necessarily represent the official views, opinions, or policy of NIH.

\section*{Data Availability}
The authors do not have permission to share patient data used in the above experiments.

\bibliographystyle{unsrt}  
\bibliography{references}   
\end{document}